%% file: main.tex
\documentclass[letterpaper]{article} 
\usepackage{times}  
\usepackage{helvet}  
\usepackage{courier}  
\usepackage{url}  
\usepackage{graphicx}  

\usepackage{times}
\usepackage{xcolor}
\usepackage{soul}
\usepackage[utf8]{inputenc}
\usepackage[small]{caption}
\usepackage{amsthm,amssymb,amsmath}
\usepackage{braket}
\usepackage{algorithm,algpseudocode}
\usepackage{comment}
\usepackage[subrefformat=parens]{subcaption}
\usepackage{url}
\usepackage{pifont}
\usepackage{booktabs}

\usepackage{tikz}
\usepackage{pgfplots}
\pgfplotsset{compat=1.14}

\makeatletter

\let\@internalcite\cite
\def\cite{\def\citeauthoryear##1##2{##1, ##2}\@internalcite}
\def\shortcite{\def\citeauthoryear##1##2{##2}\@internalcite}
\def\@biblabel#1{\def\citeauthoryear##1##2{##1, ##2}\hfill}
\makeatother

\usepackage{color}
\usepackage{CJKutf8}
\usepackage{ifthen}
\newcommand{\COMM}[2]{{
\begin{CJK}{UTF8}{ipxm}
\ifthenelse{\equal{#1}{SH}}{\color{blue}}{
\ifthenelse{\equal{#1}{TM}}{\color{red}}{
\ifthenelse{\equal{#1}{AA}}{\color{cyan}}{
\ifthenelse{\equal{#1}{BB}}{\color{magenta}}}}}
[#1: #2]
\end{CJK}
}}

\theoremstyle{definition}
\newtheorem{theorem}   {Theorem}  [section]

\newtheorem{problem}    [theorem] {Problem}

\newtheorem{example}    [theorem] {Example}

\numberwithin{equation}{section}

\newcommand{\conv}{\mathrm{conv}}
\newcommand{\ext}{\mathrm{ext}}
\DeclareMathOperator*{\argmax}{argmax}

\newcommand{\cmark}{\ding{51}}
\newcommand{\xmark}{\ding{55}}

\title{Convex Hull Approximation of Nearly Optimal Lasso Solutions}

\author{Satoshi Hara$^1$, Takanori Maehara$^2$ \\ $^1$ Osaka University \\ $^2$ RIKEN AIP \\ satohara@osaka-u.ac.jp, takanori.maehara@riken.jp}
\date{}

\begin{document}

\maketitle

\input{abstract}

\input{introduction}
\input{preliminaries}

\input{formulation}
\input{algorithm}
\input{experiments}
\input{conclusion}

\bibliographystyle{named}
\bibliography{main}

\end{document}

%% file: abstract.tex
\begin{abstract}
In an ordinary feature selection procedure, a set of important features is obtained by solving an optimization problem such as the Lasso regression problem, and we expect that the obtained features explain the data well.
In this study, instead of the single optimal solution, we consider finding a set of diverse yet nearly optimal solutions.
To this end, we formulate the problem as finding a small number of solutions such that the convex hull of these solutions approximates the set of nearly optimal solutions.
The proposed algorithm consists of two steps: 
First, we randomly sample the extreme points of the set of nearly optimal solutions.
Then, we select a small number of points using a greedy algorithm.
The experimental results indicate that the proposed algorithm can approximate the solution set well.
The results also indicate that we can obtain Lasso solutions with a large diversity.
\end{abstract}

%% file: introduction.tex
\section{Introduction}
\label{sec:introduction}

\paragraph{Background and Motivation}

\emph{Feature selection} is a procedure for finding a small set of relevant features from a dataset. 
It simplifies the model to make them easier to understand, and enhances the generalization performance; thus it plays an important role in data mining and machine learning~\cite{guyon2003introduction}.

One of the most commonly used feature selection methods is the \emph{Lasso regression}~\cite{tibshirani1996regression,chen2001atomic}.
Suppose that we have $n$ observations of $p$ dimensional vectors $x_1, \ldots, x_n \in \mathbb{R}^p$, and the corresponding responses $y_1, \ldots, y_n \in \mathbb{R}$.
Then, the Lasso regression seeks a feature vector $\beta^* \in \mathbb{R}^p$ by minimizing the $\ell_1$-penalized squared loss function 
\begin{align}
\label{eq:L}
	L(\beta) = \frac{1}{2n} \| X \beta - y \|_2^2 + \lambda \| \beta \|_1,
\end{align}
where $X = [x_1; \cdots; x_n] \in \mathbb{R}^{n \times p}$ and $y = [y_1; \cdots; y_n] \in \mathbb{R}^n$. 
Here, $\| \cdot \|_p$ denotes the $\ell_p$-norm defined by $\| \beta \|_p = (\sum_{j} |\beta_j|^p)^{1/p}$.
Since the $\ell_1$ penalty induces sparsity of the solution, we may obtain a set of features from the support of the solution.

The Lasso regression and its variants have many desirable properties; in particular, the sparsity of the solution helps users to understand which features are important for their tasks.
Hence, they are considered to be one of the most basic approaches for the cases, e.g., when models are used to support user decision making where the sparsity allows users to check whether or not the models are reliable; and, when users are interested in finding interesting mechanisms underlying the data where the sparsity enables users to identify important features and get insights of the data~\cite{guyon2003introduction}.

To further strengthen those advantages of the Lasso, Hara and Maehara~\shortcite{hara2017enumerate} proposed enumerating all (essentially different) the Lasso solutions in their increasing order of the objective values.
With the enumeration, one can find more reliable models from the enumerated solutions, or one can gain more insights of the data~\cite{hara2017enumerate,hara2018approximate}.

In this study, we aim at finding \emph{diverse} solutions instead of the exhaustive enumeration.
Hara and Maehara~\shortcite{hara2017enumerate} have observed that in real-world applications, there are too many nearly optimal solutions to enumerate them exhaustively.
Typically, if there are some highly correlated features, the enumeration algorithm outputs all their combinations as nearly optimal solutions; thus, there are exponentially many nearly optimal solutions.
Obviously, checking all those similar solutions is too exhausting for the users, which makes the existing enumeration method less practical.
To overcome this practical limitation, we consider finding diverse solutions as the representative of the nearly optimal solutions, which enables users to check ``overview'' of the solutions.

\paragraph{Contribution}

\begin{figure}[tb]
\centering
\begin{tikzpicture}[scale=0.7]
\begin{axis}[ticklabel style = {font=\Large}, label style={font=\Large}, xlabel={$\beta_1$}, ylabel={\rotatebox{270}{$\beta_2$}}]
\addplot[color={rgb:green,5;blue,1;red,1},only marks,mark=triangle*, mark size=7] coordinates {(0,1/2)};
\addlegendentry{$B(\nu^*)$}
\addplot[black,thick,dashed,samples=200,domain=0:(10+sqrt(10))/20]{(1/sqrt(10)-2*x+1)/2};
\addlegendentry{$B(\nu)$}
\addplot[blue,very thick,mark=square*, mark size=4] coordinates {(0,0.3418) (0,0.6581) (0.6581,0) (0.3418,0) (0,0.3418)};
\addlegendentry{$\conv(Q)$}
\addplot[black,thick,dashed,samples=200,domain=0:(10+sqrt(10))/20]{(1/sqrt(10)-2*x+1)/2};
\addplot[black,thick,dashed,samples=200,domain=0:(10-sqrt(10))/20]{(-1/sqrt(10)-2*x+1)/2};
\addplot[black,thick,dashed,samples=200,domain=-1/80:0]{sqrt(80*x+1)/sqrt(10)-2*x+1)/2};
\addplot[black,thick,dashed,samples=200,domain=-1/80:0]{(-sqrt(80*x+1)/sqrt(10)-2*x+1)/2};
\addplot[black,thick,dashed,samples=200,domain=(10-sqrt(10))/20:(10+sqrt(10))/20]{(3-2*x-sqrt(81-40*x-40*abs(x))/sqrt(10))/2};
\end{axis}
\end{tikzpicture}
\caption{Illustration of our approach. The triangle shows the optimal solution $B(\nu^*) = \{\beta^*\}$, the dashed line shows the boundary of the nearly optimal solutions $B(\nu)$, and the squares with the solid line show the convex hull approximation $\conv(Q)$ of $B(\nu)$.}
\label{fig:B}
\end{figure}
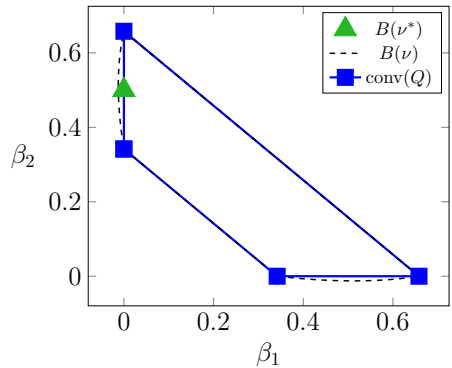

In this study, we propose a novel formulation to find diverse yet nearly optimal Lasso solutions.
Instead of the previous enumeration approach, we directly work on the set of nearly optimal solutions, defined by
\begin{align}
\label{eq:B}
	B(\nu) = \{ \beta \in \mathbb{R}^p : L(\beta) \le \nu \}
\end{align}
where $\nu \in \mathbb{R}$ is a threshold slightly greater than the optimal objective value $\nu^* = L(\beta^*)$ of the Lasso regression.

We summarize $B(\nu)$ by a small number of points $Q \subset B(\nu)$ in the sense that the convex hull of $Q$ approximates $B(\nu)$. 
We call this approach \emph{convex hull approximation}. 
Section~\ref{sec:formulation} describes the mathematical formulation of our approach.
We illustrate this approach in the following example with Figure~\ref{fig:B}.
\begin{example}
Let us consider the two-dimensional Lasso regression problem with the following loss function
\begin{align}
	L(\beta_1,\beta_2) = \frac{1}{2} \left\| \left[ \begin{matrix} 1 & 1 \\ 1 & 1 + \epsilon \end{matrix} \right] \left[ \begin{matrix} \beta_1 \\ \beta_2 \end{matrix} \right] - \left[ \begin{matrix} 1 \\ 1 \end{matrix} \right] \right\|_2^2 + \left\| \begin{matrix} \beta_1 \\ \beta_2 \end{matrix} \right\|_1
\end{align}
where $\epsilon$ is a sufficiently small parameter, e.g., $\epsilon = 1/40$. 
Then, the optimal value $\nu^*$ is approximately $3/4$, and the corresponding optimal solution $\beta^*$ is approximately $(0, 1/2)$, as shown in the green point in Figure~\ref{fig:B}.

Now, we consider the nearly optimal solution set $B(\nu)$ for the threshold $\nu = \nu^* + \epsilon$.
The boundary of this set is illustrated in the dashed line in Figure~\ref{fig:B}.
Even if $\nu - \nu^*$ is very small, since the observations $X$ is highly correlated, $B(\nu)$ contains essentially different solution, e.g., $\beta' = (1/2,0)$.

We approximate $B(\nu)$ by the convex hull of a few finite points $Q \subset B(\nu)$. 
In this case, by taking the corner points of $B(\nu)$, we can approximate this set well by the four points as shown by the blue line in Figure~\ref{fig:B}.
We note that the diversity of $Q$ is implicitly enforced because diverse $Q$ is desirable for a good approximation of $B(\nu)$; we therefore do not need to add the diversity constraint such as DPP~\cite{kulesza2012determinantal} explicitly.

This problem will be solved numerically in Section~\ref{sec:experiments}.
\qed
\label{ex:dim2}
\end{example}

We propose an algorithm to construct a good convex hull approximation of $B(\nu)$.
The algorithm consists of two steps.
First, it samples sufficiently many extreme points of $B(\nu)$ by solving Lasso regressions multiple times. 
Second, we select a small subset $Q$ from the sampled points to yield a compact summarization.
The detailed description of our algorithm is given in Section~\ref{sec:algorithm}.

We conducted numerical experiments to evaluate the effectiveness of the proposed method.
Specifically, we evaluated three aspects of the method, namely, the approximation performance, computational efficiency, and the diversity of the found solutions.
The results are shown in Section~\ref{sec:experiments}.

For simplicity, we describe only the method for Lasso regression in the manuscript; however it can be easily adopted to the other models such as the sparse logistic regression~\cite{lee2006efficient} and elastic-net~\cite{zou2005regularization}.

%% file: preliminaries.tex
\section{Preliminaries}

A set $C \subset \mathbb{R}^p$ is \emph{convex} if for any $\beta_1, \beta_2 \in C$ and $\alpha \in [0,1]$, $(1 - \alpha) \beta_1 + \alpha \beta_2 \in C$.
For a set $P \subset \mathbb{R}^p$, its \emph{convex hull}, $\conv(P)$, is the smallest convex set containing $P$.
Let $C$ be a convex set.
A point $\beta \in C$ is an \emph{extreme point} of $C$ if $\beta = (1 - \alpha) \beta_1 + \alpha \beta_2$ for some $\beta_1, \beta_2 \in C$ and $\alpha \in (0,1)$ implies $\beta = \beta_1 = \beta_2$.
The set of extreme points of $C$ is denoted by $\ext(C)$.

The Klein--Milman theorem shows the fundamental relation between the extreme points and the convex hull.
\begin{theorem}[Klein--Milman Theorem; see Barvinok~\shortcite{barvinok2002course}]
\label{thm:kleinmilman}
Let $C$ be a compact convex set. Then $\conv(\ext(C)) = C$. \qed
\end{theorem}

For two sets $C, C' \subset \mathbb{R}^p$, the \emph{Hausdorff distance} between these sets is defined by
\begin{align}
\label{eq:hausdorff_orig}
	d(C, C') = \max \{ & \sup_{\beta \in C} \inf_{\beta' \in C'} \| \beta - \beta' \|_2, \notag \\
    & \sup_{\beta' \in C'} \inf_{\beta \in C} \| \beta - \beta' \|_2 \}.
\end{align}
The Hausdorff distance forms a metric on the non-empty compact sets.
The computation of Hausdorff distance is NP-hard (more strongly, it is W[1]-hard) in general~\cite{konig2014computational}.

A function $L: \mathbb{R}^p \to \mathbb{R}$ is convex if the epigraph $\mathrm{Epi}(L) = \{ (\beta,\nu) \in \mathbb{R}^p \times \mathbb{R} : \nu \ge L(\beta) \}$ is convex.
For a convex function $L$, the level set $B(\nu) = \{ \beta \in \mathbb{R}^p : L(\beta) \le \nu \}$ is convex for all $\nu \in \mathbb{R}$.


%% file: formulation.tex
\section{Formulation}
\label{sec:formulation}

In this section, we formulate our convex hull approximation problem mathematically.
We assume that $X$ has no zero column (otherwise, we can remove the zero column and the corresponding feature from the model).

Recall that the Lasso loss function $L: \mathbb{R}^p \to \mathbb{R}$ in \eqref{eq:L} is convex; therefore, the set of nearly optimal solutions $B(\nu)$ in \eqref{eq:B} forms a closed convex set.
Moreover, since $X$ has no zero column, $B(\nu)$ is compact.

Our goal is to summarize $B(\nu)$.
By the Klein--Milman theorem (Theorem~\ref{thm:kleinmilman}), $B(\nu)$ can be reconstructed from the extreme points of $B(\nu)$ as $B(\nu) = \conv(\ext(B(\nu)))$;  therefore, it is natural to output the extreme points $\ext(B(\nu))$ as a summary of $B(\nu)$.
If $\nu = \nu^*$, this approach corresponds to enumerating the vertices of $B(\nu^*)$, which forms a polyhedron~\cite{tibshirani2013lasso}; therefore, we can use the existing algorithm to enumerate the vertices of a polyhedron developed in Computational Geometry~\cite{fukuda1997analysis} as in Pantazis \textit{et al.}~\shortcite{pantazis2017enumerating}.
However, if $\nu > \nu^*$, $B(\nu)$ has a piecewise smooth boundary, as shown in \figurename~\ref{fig:B}; therefore, there are continuously many extreme points of $B(\nu)$, which cannot be enumerated.\footnote{Pantazis \textit{et al.}~\shortcite{pantazis2017enumerating} also consider the near optimal solutions. 
However, they focused only on the subset of $B(\nu)$ spanned by the support of the Lasso global solution.
We do not take this approach since it cannot handle a global structure of $B(\nu)$.}

Therefore, we select a finite number of points $Q \subset \ext(B(\nu))$ as a ``representative'' of the extreme points such that $\conv(Q)$ well approximates $B(\nu)$.
We measure the quality of the approximation by the Hausdorff distance~\eqref{eq:hausdorff_orig}.

To summarize the above discussion, we pose the following problem.
\begin{problem}
We are given a loss function $L: \mathbb{R}^p \to \mathbb{R}$ and a threshold $\nu \in \mathbb{R}$. Let $B(\nu) = \{ \beta \in \mathbb{R}^p : L(\beta) \le \nu \}$. 
Find a point set $Q \subset B(\nu)$ such that (1) $d(\conv(Q), B(\nu))$ is small, and (2) $|Q|$ is small.
\end{problem}

The problem of approximating a convex set by a polyhedron has a long history in convex geometry (see Bronstein~\shortcite{bronstein2008approximation} for a recent survey).
Asymptotically, for any compact convex set with a smooth boundary, the required number of points to obtain an $\epsilon$ approximation is $\Theta(\sqrt{p}/\epsilon^{(p-1)/2})$~\cite{bronstein1975approximation,gruber1993aspects}.
Therefore, in the worst case, we may need exponentially many points to have a reasonable approximation.

On the other hand, if we focus on the non-asymptotic $\epsilon$, we have a chance to obtain a simple representation.
One intuitive situation is that the polytope $B(\nu^*)$ has a small number of vertices, as in \figurename~\ref{fig:B}. 
In such a case, by taking the vertices as $Q$, we can obtain an $O(\epsilon)$ approximation for $B(\nu)$ when $\nu = \nu^* + O(\epsilon)$.

Therefore, below, we assume that $B(\nu)$ admits a small number of representatives and construct an algorithm to find such representatives.

%% file: algorithm.tex
\section{Algorithm}
\label{sec:algorithm}

In this section, we propose a method to compute a convex hull approximation $\conv(Q)$ of $B(\nu)$.

Since $\conv(Q) \subseteq B(\nu)$ and the sets are compact, the Hausdorff distance between $\conv(Q)$ and $B(\nu)$ is given by
\begin{align}
\label{eq:hausdorff}
	d(\conv(Q), B(\nu)) = \max_{\beta \in B(\nu)} \min_{\beta' \in \conv(Q)} \| \beta - \beta' \|_2.
\end{align}
A natural approach is to minimize this quantity by a greedy algorithm that successively selects the maximizer $\beta \in B(\nu)$ of \eqref{eq:hausdorff} and then adds it to $Q$.
However, this approach is impractical, because the optimization problem \eqref{eq:hausdorff} is a convex \emph{maximization} problem.\footnote{In our preliminary study, we implemented the projected gradient method to find the farthest point $\beta \in B(\nu)$. However, it was slow, and often converged to poor local maximal solutions.}

To overcome this difficulty, we use a random sampling approximation for $B(\nu)$.
We first sample sufficiently many points $S$ from $\ext(B(\nu))$ and then regard $\conv(S)$ as an approximation of $B(\nu)$.
Once this approximation is constructed, the maximum in \eqref{eq:hausdorff} can be obtained by a simple linear search.
Therefore, this reduces our problem to a simple subset selection problem.

The overall procedure of our algorithm is shown in Algorithm~\ref{alg:overview}. 
It consists of two steps: random sampling step and subset selection step.
Below, we describe each step.

\begin{algorithm}[tb]
\begin{algorithmic}[1]
\State{Sample sufficiently many points $S \subset B(\nu)$ by Algorithm~\ref{alg:sampler}.}
\State{Select a few points $Q \subseteq S$ by Algorithm~\ref{alg:selecter}.}
\end{algorithmic}
\caption{Proposed algorithm}
\label{alg:overview}
\end{algorithm}

\subsection{Sampling Extreme Points (Algorithm~\ref{alg:sampler})}

\figurename~\ref{fig:B} suggests that selecting corner points of $B(\nu)$ as $Q$ is desirable to obtain a good approximation of $B(\nu)$.
Here, to obtain a good candidate of $Q$, we consider a sampling algorithm that samples the corner points of $B(\nu)$. 

First, we select a uniformly random direction $d \in \mathbb{R}^p$. 
Then, we find the extreme point $\beta \in B(\nu)$ by solving the following problem
\begin{align}
\label{eq:primal}
	\max \{ d^\top \beta : \beta \in B(\nu) \}.
\end{align}
We solve this problem by using the Lagrange dual with binary search as follows.
With the Lagrange duality, we obtain the following equivalent problem
\begin{align}
\label{eq:dual}
	\min_{\tau \ge 0} D(\tau|d) := \max_{\beta} \left( d^\top \beta - \tau (L(\beta) - \nu) \right) .
\end{align}
Since the optimal solution $\beta(d)$ of \eqref{eq:primal} satisfies $L(\beta(d)) = \nu$, we seek $\tau$ by using a binary search\footnote{Since we have no upper bound of the search range, we actually use the exponential search that successively doubles the search range~\cite{bentley1976almost}.}
so that $L(\beta(d)) = \nu$ to hold.
%
%

It should be noted that the proposed sampling algorithm can be completely parallelized.

\medskip 

\noindent
\textbf{Properties of Sampling}\,
The solution to the problem \eqref{eq:primal} tends to be sparse because of the $\ell_1$ term in $L(\beta)$, which indicates that we can sample a corner point of $B(\nu)$ in the direction of $d$, such as the ones in \figurename~\ref{fig:B}.
More precisely, the proposed algorithm samples each extreme point with probability proportional to the volume of the normal cone of each point.
Because the corner points have positive volumes, the algorithm samples corner points with high probabilities.

\begin{algorithm}[tb]
\begin{algorithmic}[1]
\State{$S = \emptyset$}
\For{$j=1, 2, \ldots, M$}
\State{Draw $d \sim \mathcal{N}(0, I)$}
\State{Solve \eqref{eq:primal} to obtain an extreme point $\beta(d)$ and add it to $S$}
\EndFor
\State{Return $S$}
\end{algorithmic}
\caption{Sampling points}
\label{alg:sampler}
\end{algorithm}

\subsection{Greedy Subset Selection (Algorithm~\ref{alg:selecter})}

\begin{algorithm}[tb]
\begin{algorithmic}[1]
\State{Select $\beta_1 \in S$ arbitrary and let $Q = \{\beta_1\}$}
\State{Initialize a heap data structure $H$ by $H[\beta] \leftarrow d(\beta, Q)$ for all $\beta \in S \setminus Q$}
\While{$|Q| < K$}
\State{Let $\beta \in H$ be the point that has the largest $H[\beta]$}
\State{Update $H[\beta] \leftarrow d(\beta, Q)$}
\If{$H[\beta]$ is still the largest point}
\State{Add $\beta$ to $Q$ and remove $\beta$ from the heap}
\EndIf
\EndWhile
\State{Output $Q$}
\end{algorithmic}
\caption{Select points}
\label{alg:selecter}
\end{algorithm}

Next, we select a small subset $Q \subseteq S$ from the sampled points $S \subset \mathbb{R}^p$ that do not lose the approximation quality.

We use the farthest point selection method, proposed in Blum~\textit{et al.}~\shortcite{blum2016sparse}.\footnote{Blum~\textit{et al.}~\shortcite{blum2016sparse} called this procedure \emph{Greedy Clustering}.}
In this procedure, we start from any point $\beta_1 \in S$. 
Then, we iteratively select the point $\beta_j \in S$ by solving the sample-approximated version of \eqref{eq:hausdorff}, i.e., the farthest point from the convex hull is taken as
\begin{align}
	\beta_j \in \argmax_{\beta \in S} \min_{\beta' \in \conv(\{\beta_1, \ldots, \beta_{j-1}\})} \| \beta - \beta' \|_2.
\end{align}

This procedure has the following theoretical guarantee.
\begin{theorem}[Blum~\textit{et al.}~\shortcite{blum2016sparse}]
Let $S \subset \mathbb{R}^p$ be a finite set enclosed in the unit ball.
Suppose that there exists a finite set $Q^* \subset S$ of size $k^*$ such that $d(\conv(Q^*), \conv(S)) \le \epsilon$.
Then, the greedy algorithm finds a set $Q \subset S$ of size $k^*/\epsilon^{2/3}$ with $d(\conv(Q), \conv(S)) = O(\epsilon^{1/3})$. 
\qed
\end{theorem}
Thus, if the number of samples $|S|$ are sufficiently large such that $d(\conv(S), B(\nu)) \le \epsilon$, the algorithm finds a convex hull approximation with $O(\epsilon^{1/3})$ error.

Below, we describe how to implement this procedure.
First, the distance from $\beta$ to the convex hull of $\beta_1, \ldots, \beta_k$ is computed by solving the following problem:
\begin{align}
\label{eq:distance}
    \begin{split}
    \min_\alpha & \textstyle{\| \beta - \sum_j \alpha_j \beta_j \|^2}  \\
    \text{s.t.} & \; \textstyle{\sum_j \alpha_j = 1, \; \alpha_j \ge 0.}
    \end{split}
\end{align}
This problem is a convex quadratic programming problem, which can be solved efficiently by using the interior point method~\cite{achache2006new}.

To implement the greedy algorithm, we have to evaluate the distance from each point to the current convex hull.
However, this procedure can be expensive when $|S|$ is large as we need to solve the problem (\ref{eq:distance}) many times.
For efficient computation, we need to avoid evaluating the distance as much as possible.

We observe that, if we add a new point $\beta_j$ to the current convex hull, the distances from other points to the convex hull decrease monotonically.
Therefore, we can use the \emph{lazy update technique}~\cite{minoux1978accelerated} to accelerate the procedure as follows.

We maintain the points $S$ by a heap data structure whose keys are the upper bounds of the distance to the convex hull.
First, we select an arbitrary point $\beta_1$, and then initialize the key of $\beta \in S$ by $d(\beta_1, \beta)$.
For each step, we select the point $\beta_j' \in S$ from the heap such that $\beta_j'$ has the largest distance upper bound. 
Then, we recompute the distance $d(\beta_j', \conv(Q))$ by solving the quadratic program~\eqref{eq:distance} and update the key of $\beta_j'$.
If it still has the largest distance upper bound, it is the farthest point; therefore, we select $\beta_j'$ as the $j$-th point $\beta_j$.
Otherwise, we repeat this procedure until we find the farthest point.
See Algorithm~\ref{alg:selecter} for the detail.

%% file: experiments.tex
\section{Experiments}
\label{sec:experiments}

We evaluate the three aspects of the proposed algorithm, namely, the approximation performance, computational efficiency, and the diversity of the found solutions.
First, we visualize the results of the algorithm by using a low dimensional synthetic data (Section~\ref{sec:visualdemonstration}).
Then, we evaluate the approximation performance and computational efficiency by using a larger dimensional synthetic data (Sections~\ref{sec:approximationperformance} and \ref{sec:computationalefficiency}).
Finally, we evaluate the diversity of the obtained solutions by using real-world datasets (Section~\ref{sec:diversity} and \ref{sec:results_other}).

\paragraph{Sample Approximation of  Hausdorff Distance for Evaluation}

We evaluate the approximation performance by the Hausdorff distance between the obtained convex hull and $B(\nu)$.
However, the exact Hausdorff distance cannot be computed since it requires solving a convex maximization problem.
We therefore adopt the sample approximation of Hausdorff distance, which is derived as follows:
\begin{enumerate}
	\item Sample $M'$ extreme points $Q^*$ by using Algorithm~\ref{alg:sampler}.
	\item Define the sample approximation of $B(\nu)$ by the convex hull $\conv(Q^*)$.
	\item Measure the Hausdorff distance $d(\conv(Q), \conv(Q^*))$ as an approximation of $d(\conv(Q), B(\nu))$.
\end{enumerate}

\paragraph{Implementations}
The codes were implemented in Python 3.6.
In Algorithm~\ref{alg:sampler}, to solve the problem (\ref{eq:dual}), we used \texttt{enet\_coordinate\_descent\_gram} function in \texttt{scikit-learn}.
In Algorithm~\ref{alg:selecter}, we selected the first point $\beta_1 \in S$ as $\beta_1 =\argmax_{\beta \in S} \|\beta - \beta^*\|_2$.
To compute the projection (\ref{eq:distance}), we used \texttt{CVXOPT} library.
The experiments were conducted on a system with an Intel Xeon E5-1650 3.6GHz CPU and 64GB RAM, running 64-bit Ubuntu 16.04.

\subsection{Visual Demonstration}
\label{sec:visualdemonstration}

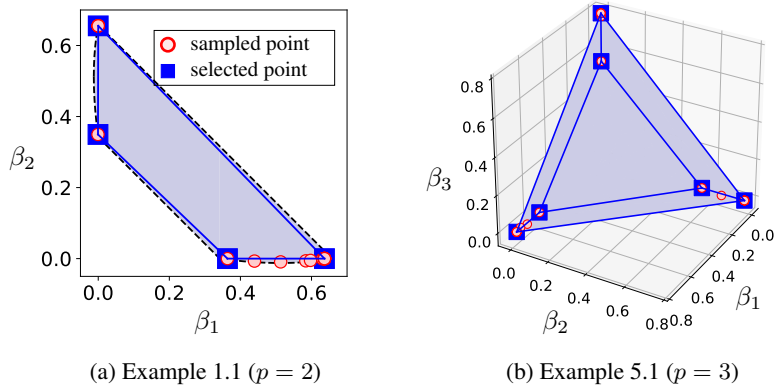
\begin{figure}[t]
	\centering
	\begin{minipage}[b]{0.45\textwidth}
	\begin{tikzpicture}
		\node[above right] (img) at (0,0) {\includegraphics[width=0.8\textwidth]{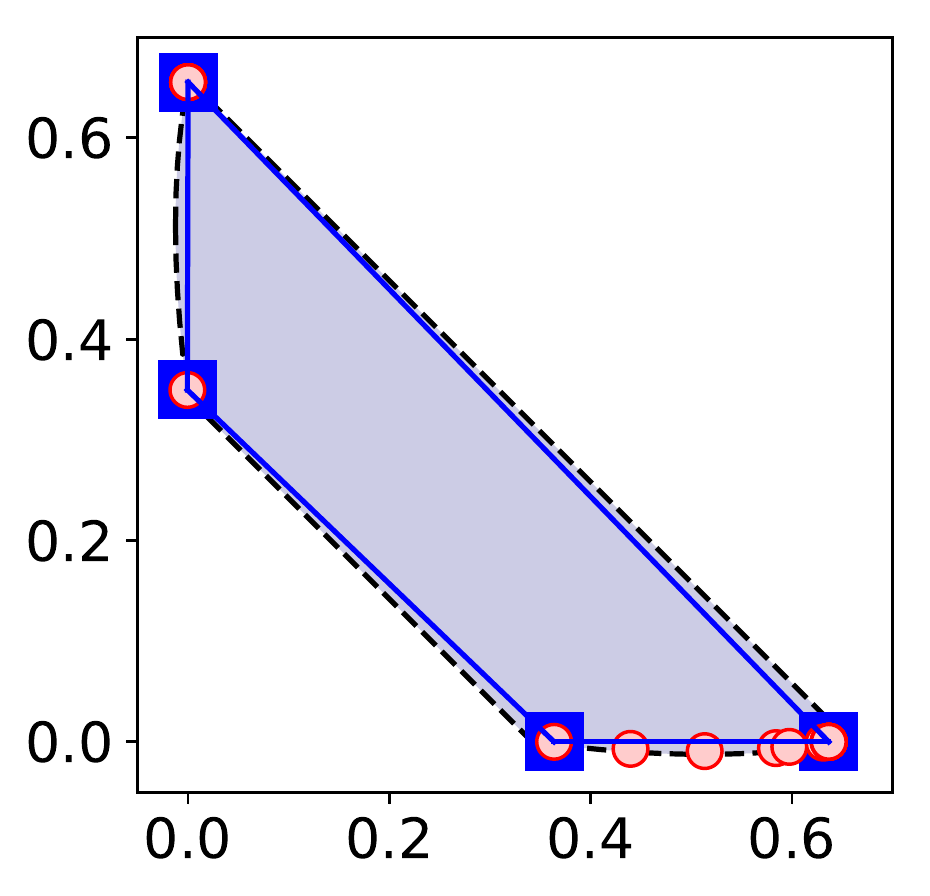}};
		\node at (70pt, 0pt) {$\beta_1$};
		\node at (0pt, 62pt) {$\beta_2$};
		\draw [very thin] (1.75, 3.15) rectangle (4.15, 3.88);
		\node[circle, draw=red, fill={rgb:red,1;white,10}, inner sep=0pt, minimum size=5pt, line width=1pt] (a) at (55pt, 105pt) {};
		\node[rectangle, draw=blue, fill=blue, inner sep=0pt, minimum size=5pt] (b) at (55pt, 95pt) {};
		\node[right] at (60pt, 105pt) {\footnotesize sampled point};
		\node[right] at (60pt, 95pt) {\footnotesize selected point};
	\end{tikzpicture}
	\subcaption{Example~\ref{ex:dim2} ($p=2$)}
	\label{fig:dim2_result}
	\end{minipage}
	\begin{minipage}[b]{0.45\textwidth}
	\begin{tikzpicture}
		\node[above right] (img) at (0,0) {\includegraphics[width=0.9\textwidth]{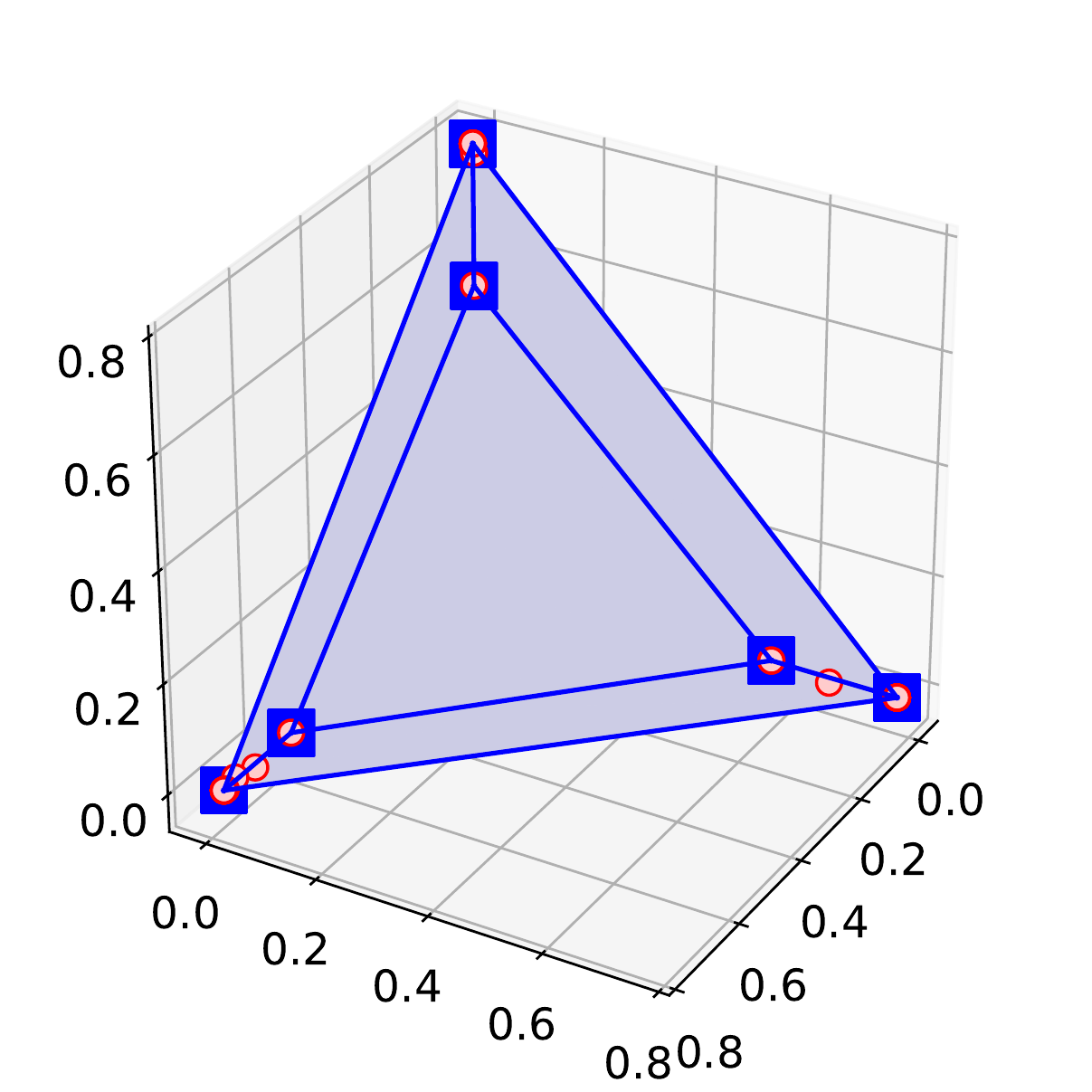}};
		\node at (120pt, 17pt) {$\beta_1$};
		\node at (47pt, 8pt) {$\beta_2$};
		\node at (3pt, 62pt) {$\beta_3$};
	\end{tikzpicture}
	\subcaption{Example~\ref{ex:dim3} ($p=3$)}
	\label{fig:dim3_result}
	\end{minipage}
	\caption{Visual demonstrations of the proposed method in low-dimensions - (a) Result for Example \ref{ex:dim2}, where $B(\nu)$ is indicated by the shaded regions, (b) Results for Example \ref{ex:dim3}.}
	\label{fig:lowdim}
\end{figure}

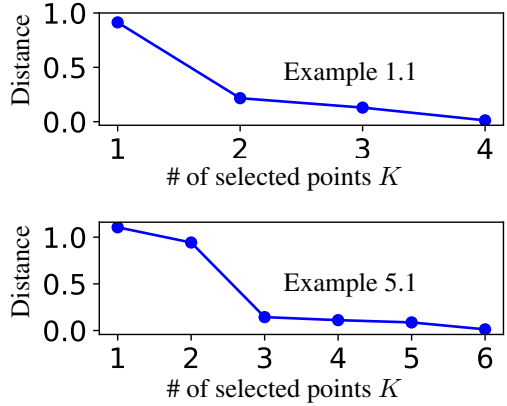
\begin{figure}[t]
	\centering
	\begin{minipage}[b]{0.6\textwidth}
	\begin{tikzpicture}
		\node[above right] (img) at (0,0) {\includegraphics[width=0.9\textwidth]{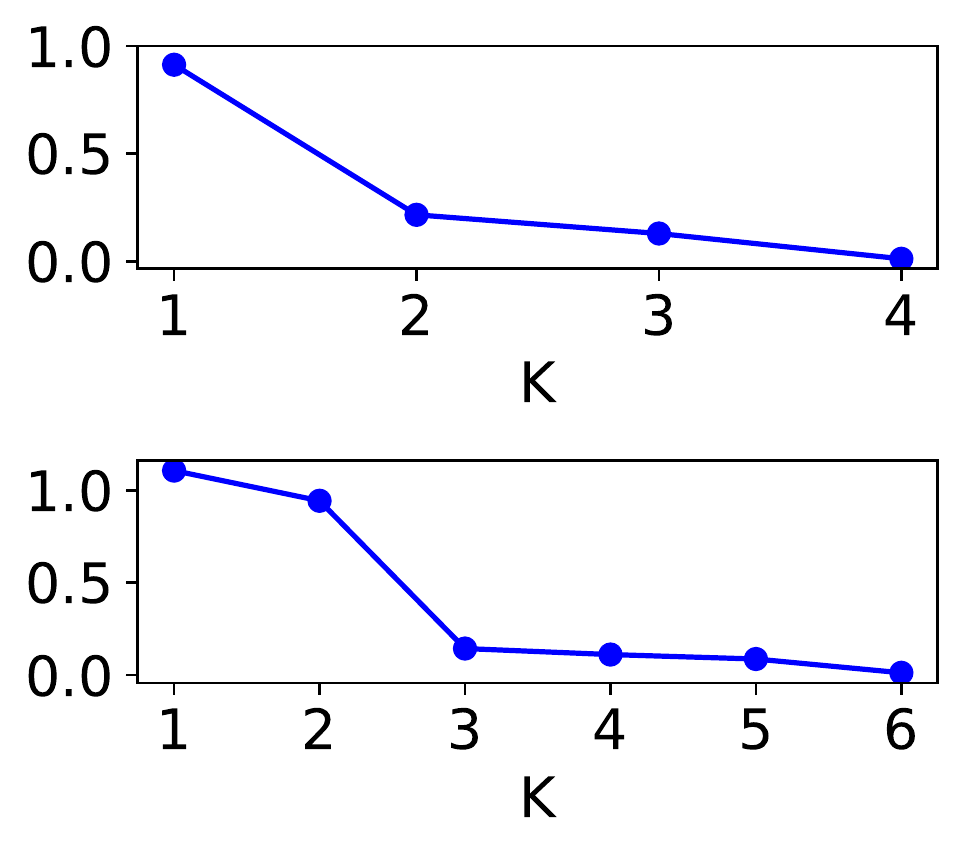}};
		\node[fill={rgb:white,1}] at (100pt, 95pt) {\# of selected points $K$};
		\node[fill={rgb:white,1}] at (100pt, 15pt) {\# of selected points $K$};
		\node at (125pt, 135pt) {Example~\ref{ex:dim2}};
		\node at (125pt, 55pt) {Example~\ref{ex:dim3}};
		\node at (0pt, 140pt) {\rotatebox{90}{Distance}};
		\node at (0pt, 60pt) {\rotatebox{90}{Distance}};
	\end{tikzpicture}
    \end{minipage}
	\caption{Approximation errors of $B(\nu)$, measured by the Hausdorff distance ($M'=1,000$).}
	\label{fig:lowdim2}
\end{figure}

For visual demonstration, we consider two examples: Example~\ref{ex:dim2} in Section~1 and Example~\ref{ex:dim3} defined below.
\begin{example}
Consider the three-dimensional Lasso regression problem with the following loss function
\begin{align}
	L(\beta) = \frac{1}{2} \left\| \left[ \begin{matrix} 1 & 1 & 1 \\ 1 & 1 + \epsilon & 1 \\ 1 & 1 & 1 + 2 \epsilon \end{matrix} \right] \left[ \begin{matrix} \beta_1 \\ \beta_2 \\ \beta_3 \end{matrix} \right] - \left[ \begin{matrix} 1 \\ 1 \\ 1 \end{matrix} \right] \right\|_2^2 + \left\| \begin{matrix} \beta_1 \\ \beta_2 \\ \beta_3 \end{matrix} \right\|_1
\end{align}
where $\epsilon$ is a sufficiently small parameter.
Then, the optimal value $\nu^*$ is approximately $5/6$, and the corresponding optimal solution $\beta^*$ is approximately $(0, 0, 2/3)$.
We set $\epsilon=1/40$, and define the set $B(\nu)$ by $\nu = 5/6 + \epsilon$.
\label{ex:dim3}
\end{example}

In Example~\ref{ex:dim2}, because of the correlation between the two features, there exists a nearly optimal solution $\beta = (1/2, 0)$ apart from the optimal solution $\beta^* = (0, 1/2)$.
The objective of the proposed method is therefore to find a convex hull that covers these solutions.
Similarly, in Example~\ref{ex:dim3}, three features are highly correlated.
The objective is to find a convex hull that covers nearly optimal solutions such as $\beta = (2/3, 0, 0)$, $(0, 2/3, 0)$, and $(0, 0, 2/3)$.

\figurename~\ref{fig:lowdim} and \ref{fig:lowdim2} show the results of the proposed method for the examples.
Here, we set the number of samples $M$ in Algorithm~\ref{alg:sampler} to be 50, and the number of greedy point selection selection $K=|Q|$ in Algorithm~\ref{alg:selecter} to be four and six, respectively.
Figures~\ref{fig:lowdim}(a) and \ref{fig:lowdim}(b) show that the proposed method successfully approximated $B(\nu)$ by using a few points. 
Indeed, as shown in \figurename~\ref{fig:lowdim2}, the approximation errors converged to almost zeros indicating that $B(\nu)$ is well-approximated with convex hulls.

\subsection{Approximation Performance}
\label{sec:approximationperformance}

We now turn to exhaustive experiments to verify the performance of the proposed algorithm in general settings.
Specifically, we show that the proposed algorithm can approximate $B(\nu)$ well, even in higher dimensions.

We generate higher dimensional data by
\begin{align}
\label{eq:mid_data}
	x \sim \mathcal{N}(0_p, \Sigma), \;\;\; y = x^\top \beta + \varepsilon, \;\;\; \varepsilon \sim \mathcal{N}(0, 0.01),
\end{align}
where $\Sigma_{ij} = \exp(- 0.1 |i - j|)$, and $\beta_i = 10/p$ if $\mod(i-1, 10)=0$ and $\beta_i = 0$ otherwise.
Because of the correlations induced by $\Sigma$, the neighboring features in $x$ are highly correlated, which indicates that there may exist several nearly optimal $\beta$. 

We set the number of observations $n$ to be $p=2n$, and the regularization parameter $\lambda$ to be 0.1.
We also define the set $B(\nu)$ by setting $\nu = 1.01 L(\beta^*)$.

\medskip 

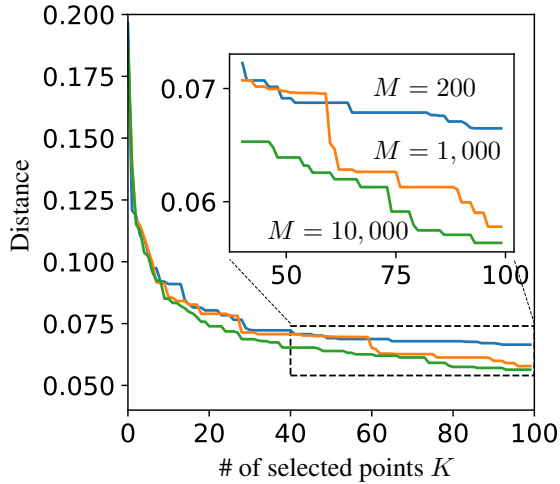
\begin{figure}[t]
	\centering
    \begin{minipage}[t]{0.7\textwidth}
	\begin{tikzpicture}
		\node[above right] (img) at (0,0) {\includegraphics[width=0.85\textwidth]{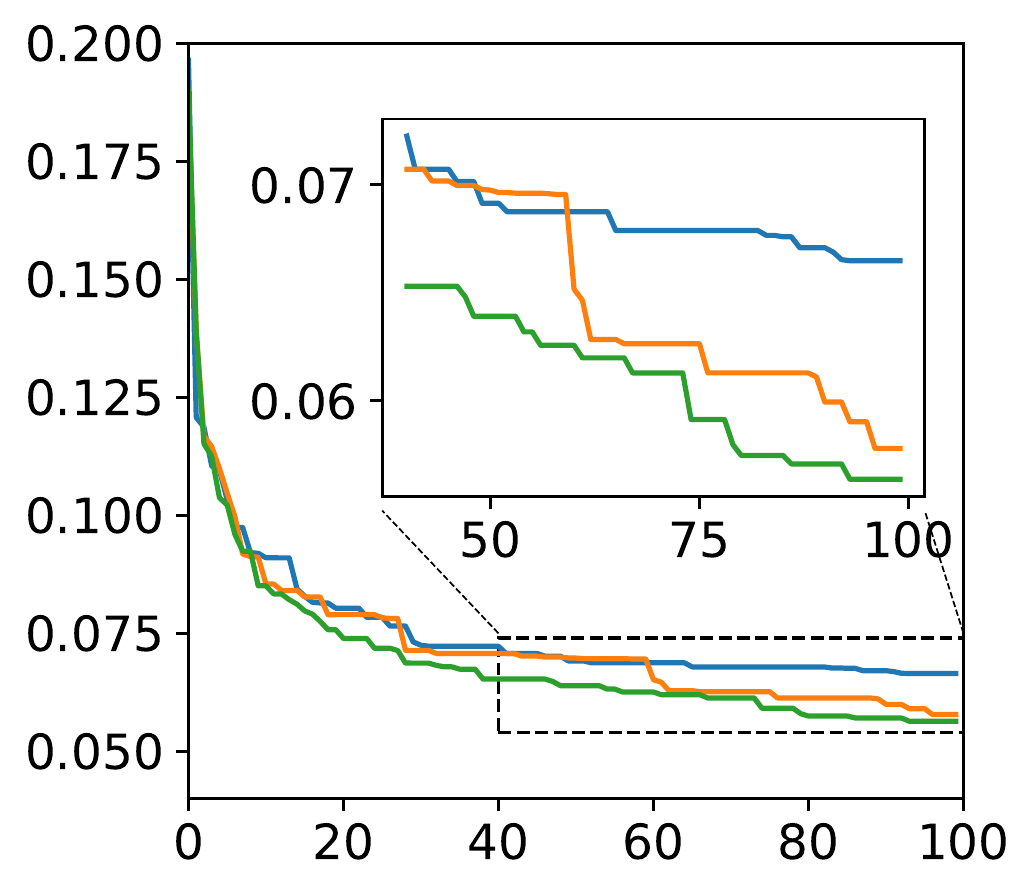}};
		\node[right] at (130pt, 145pt) {$M=200$};
		\node[right] at (130pt, 120pt) {$M=1,000$};
		\node[right] at (90pt, 90pt) {$M=10,000$};
		\node at (120pt, 0pt) {\# of selected points $K$};
		\node at (0pt, 100pt) {\rotatebox{90}{Distance}};
	\end{tikzpicture}
	\caption{\# of selected points $K$ vs.\ Hausdorff distance ($M'=100,000$)}
	\label{fig:middim}
	\end{minipage}
\end{figure}

\figurename~\ref{fig:middim} is the result for $p=100$.
The figure shows that the proposed algorithm can approximate $B(\nu)$ well.
In the figure, there are two important observations.
First, as the number of samplings $M$ increases, the Hausdorff distance decreases, indicating that the approximation performance improves.
This result is intuitive in that a many greater number of candidate points lead to a better approximation.

Second, the choice of the number of samplings $M$ is not that fatal in practice.
The result shows that the difference between the Hausdorff distances for $M=200$ and for $M=10,000$ is subtle.
It also shows that the the Hausdorff distance for $M=1,000$ and for $M=10,000$ are almost identical for larger $K$.
This indicates that we do not have to sample many points in practice.

\begin{figure}[t]
	\centering
    \begin{minipage}[t]{0.8\textwidth}
	\begin{tikzpicture}
		\node[above right] (img) at (0,0) {\includegraphics[width=0.98\textwidth]{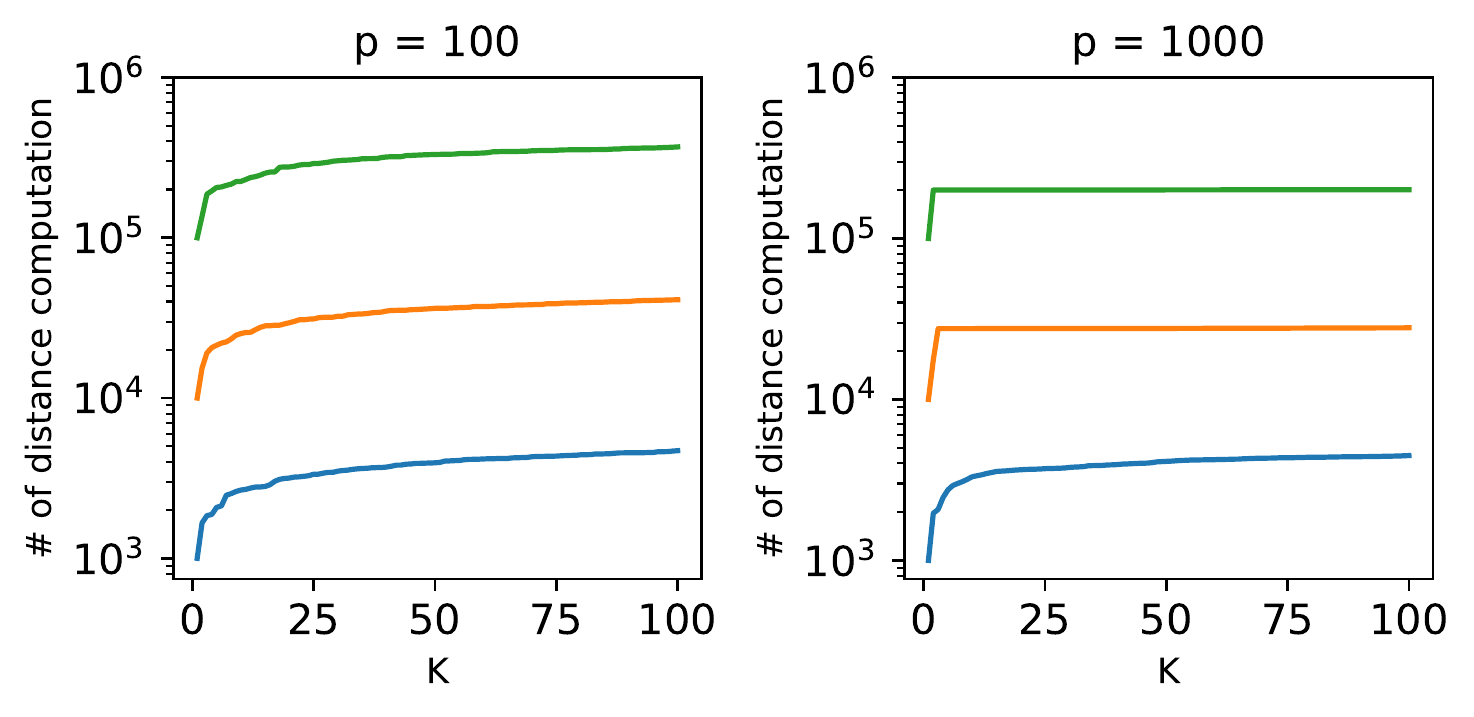}};
		\node[fill=white] at (86pt, 8pt) {\footnotesize \# of selected points $K$};
		\node[fill=white] at (222pt, 8pt) {\footnotesize \# of selected points $K$};
		\node[fill=white] at (7pt, 75pt) {\footnotesize \rotatebox{90}{\# of distance evaluation}};
		\node[fill=white] at (145pt, 75pt) {\footnotesize \rotatebox{90}{\# of distance evaluation}};
		\node[fill=white] at (85pt, 130pt) {\footnotesize $p=100$};
		\node[fill=white] at (220pt, 130pt) {\footnotesize $p=1,000$};
		\node[right] at (50pt, 42pt) {\scriptsize $M=1,000$};
		\node[right] at (50pt, 70pt) {\scriptsize $M=10,000$};
		\node[right] at (50pt, 98pt) {\scriptsize $M=100,000$};
		\node[right] at (187pt, 44pt) {\scriptsize $M=1,000$};
		\node[right] at (187pt, 70pt) {\scriptsize $M=10,000$};
		\node[right] at (187pt, 96pt) {\scriptsize $M=100,000$};
	\end{tikzpicture}
	\caption{\# of distance evaluations in Algorithm~\ref{alg:selecter}}
	\label{fig:mid_count}
	\end{minipage}
\end{figure}

\subsection{Computational Efficiency}
\label{sec:computationalefficiency}

Next, we evaluate the computational efficiency by using the same setting \eqref{eq:mid_data} used in the previous section.

\tablename~\ref{tab:middim_time} shows the runtimes of the proposed method for $p=100$ and $p=1,000$ by fixing $K = 100$.
The computational time for the sampling step increases as the number of samples $M$ and the dimension $p$ increases.
Since the approximation performance does not improve so much as the number of samples $M$ increases as observed in the previous experiment, it is helpful in practice to use a moderate number of samples.
The computational time for the greedy selection step also increases as the number of samples $M$ increases; however, interestingly, it \emph{decreases} as the dimension $p$ increases. 
This reason is understood by observing the number of distance evaluations as follows.

\begin{table}[t]
	\centering
	\caption{Runtime (in sec.) of the proposed algorithm (Sampling: Algorithm~\ref{alg:sampler}, Greedy: Algorithm~\ref{alg:selecter}) for selecting $K=100$ vertices over the numbers of samplings $M = 1,000, 10,000$, and $100,000$.}
	\label{tab:middim_time}
	\begin{tabular}{ccccc}
        \toprule
		& \multicolumn{2}{c}{$p=100$} & \multicolumn{2}{c}{$p=1,000$} \\
		$M$ & Sampling & Greedy & Sampling & Greedy \\ \midrule
		1,000 & 2.891 & 34.87 & 46.54 & 17.99 \\
		10,000 & 27.80 & 178.5 & 2466 & 66.95 \\
		100,000 & 279.1 & 1548 & 4586 & 379.9 \\
        \bottomrule
	\end{tabular}
\end{table}

\figurename~\ref{fig:mid_count} shows the number of distance evaluations in the greedy selection step in each $K$. 
This shows that the redundant distance computation is significantly reduced by the lazy update technique; therefore Algorithm~\ref{alg:selecter} only performs a few distance evaluation in each iteration.
In particular, for $p = 1,000$, the saturation is very sharp; thus the computational cost is significantly reduced.
This may be because, in high-dimensional problems, adding one point to the current convex hull does not change the distance to the remaining points, and hence the lazy update helps to avoid most of distance evaluations.

\subsection{Diversity of Solutions}
\label{sec:diversity}

One of the practical advantages of the proposed method is that it can find nearly optimal solutions with a large diversity.
This is a favorable property when one is interested in finding several possible explanations for a given data, which is usually the case in data mining.

\paragraph{Setup}
Here, we verify the diversity of the found solutions on the 20 Newsgroups data.\footnote{\url{http://qwone.com/~jason/20Newsgroups/}}
The results on other datasets can be found in the next section.
In this experiment, we consider classifying the documents between the two categories \texttt{ibm.pc.hardware} and \texttt{mac.hardware}.
As a feature vector $x$, we used tf-idf weighted bag-of-words expression, with stop words removed.
The dataset comprised $n = 1,168$ samples with $p = 11,648$ words.
Our objective is to find discriminative words that are relevant to the classification of the documents.

Because the task is binary classification with $y \in \{-1, +1\}$, instead of the squared objective, we use the Lasso logistic regression with the objective function given as
\begin{align}
\label{eq:logreg}
	L(\beta) = \frac{1}{n} \sum_{i=1}^n \log \left(1 + \exp(-y_ix_i^\top \beta)\right) + \lambda \|\beta\|_1 .
\end{align}
We implemented the solver for the problem (\ref{eq:dual}) by modifying \texttt{liblinear}~\cite{REF08a}.
In the experiment, we set the regularization parameter $\lambda$ to be $0.001$.

\paragraph{Baseline Methods}
We compared the solution diversity of the proposed method with the two baselines in Hara and Maehara~\shortcite{hara2017enumerate}.
The first baseline simply enumerates the optimal solutions with different supports in the ascending order of the objective function value~(\ref{eq:logreg}).
We refer to this method as \textit{Enumeration}.
The second baseline employs a heuristics to skip similar solutions during the enumeration.
It can therefore improve the diversity of the enumerated solutions.
We refer to this heuristic method as \textit{Heuristic}.
Note that we did not adopt the method of Pantazis \textit{et al.}~\shortcite{pantazis2017enumerating} as the baseline because it enumerates only the sub-support of the Lasso global solution: it cannot find solutions apart from the global solution.

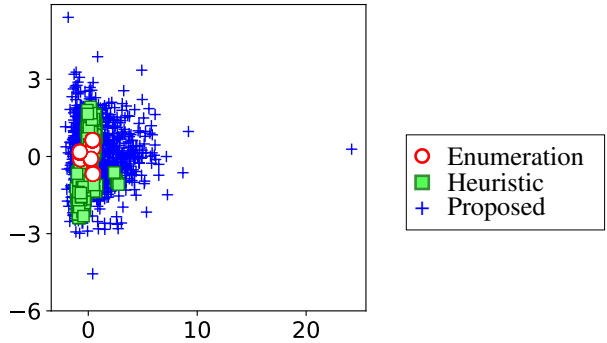
\begin{figure}[t]
	\vspace{-16pt}
	\centering
    \begin{minipage}[t]{0.7\textwidth}
    \centering
	\begin{tikzpicture}
		\node[above right] (img) at (-125pt,0pt) {\includegraphics[width=0.6\textwidth]{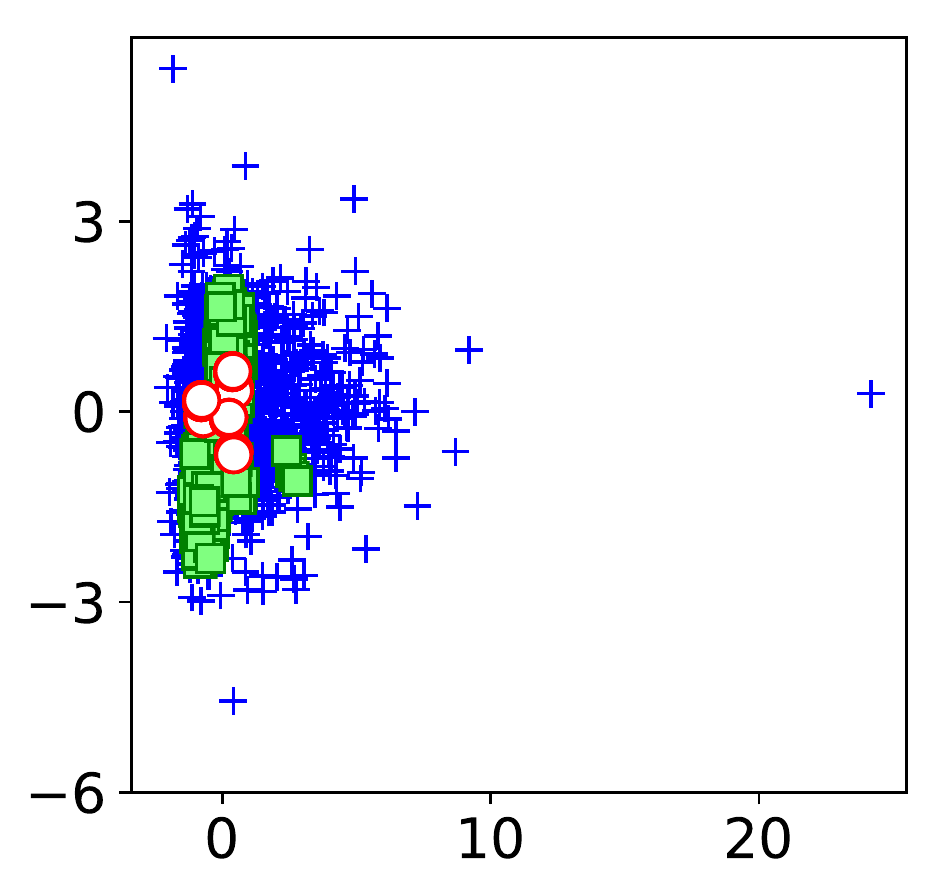}};
        \draw [very thin] (33pt, 50pt) rectangle (108pt, 86pt);
		\node[circle, draw=red, fill=white, inner sep=0pt, minimum size=5pt, line width=1pt] (a) at (39pt, 78pt) {};
		\node[rectangle, draw={rgb:green,5;blue,2;red,2}, fill={rgb:green,2;white,1}, inner sep=0pt, minimum size=5pt, line width=1pt] (b) at (39pt, 68pt) {};
		\node[text=blue] (c) at (39pt, 58pt) {+};
		\node[right] at (45pt, 78pt) {Enumeration};
		\node[right] at (45pt, 68pt) {Heuristic};
		\node[right] at (45pt, 58pt) {Proposed};
	\end{tikzpicture}
	\caption{Found 500 solutions in 20 Newsgroups data, shown in 2D using PCA.}
	\label{fig:20news_diversity}
    \end{minipage}
\end{figure}

\begin{table}[t]
	\centering
	\caption{Representative words found in 20 Newsgroups data}
	\label{tab:20news_words}
	\begin{tabular}{c ccc}
    \toprule
		& Enumeration & Heuristic & Proposed  \\
        \midrule
		``apple'' & \textcolor{blue}{\cmark} & \textcolor{blue}{\cmark} & \textcolor{blue}{\cmark} \\
		``macs'' & \textcolor{red}{\xmark} & \textcolor{blue}{\cmark} & \textcolor{blue}{\cmark} \\
		``macintosh'' & \textcolor{red}{\xmark} & \textcolor{red}{\xmark} & \textcolor{blue}{\cmark} \\
        \bottomrule
	\end{tabular}
\end{table}

\begin{figure*}[t]
	\begin{minipage}[t]{0.26\textwidth}
	\includegraphics[width=0.93\textwidth]{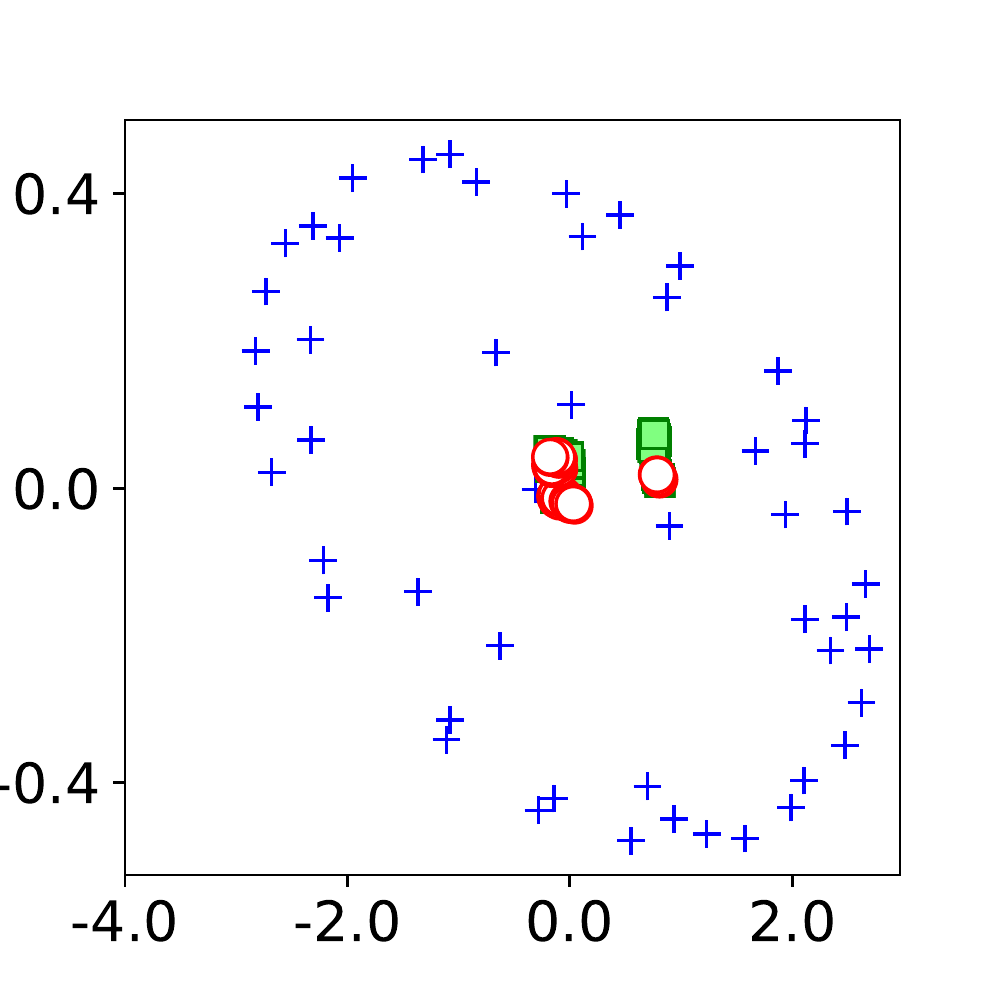}
	\subcaption{cputime}
	\label{fig:cpusmall_diversity}
	\end{minipage}
	\begin{minipage}[t]{0.26\textwidth}
	\includegraphics[width=0.93\textwidth]{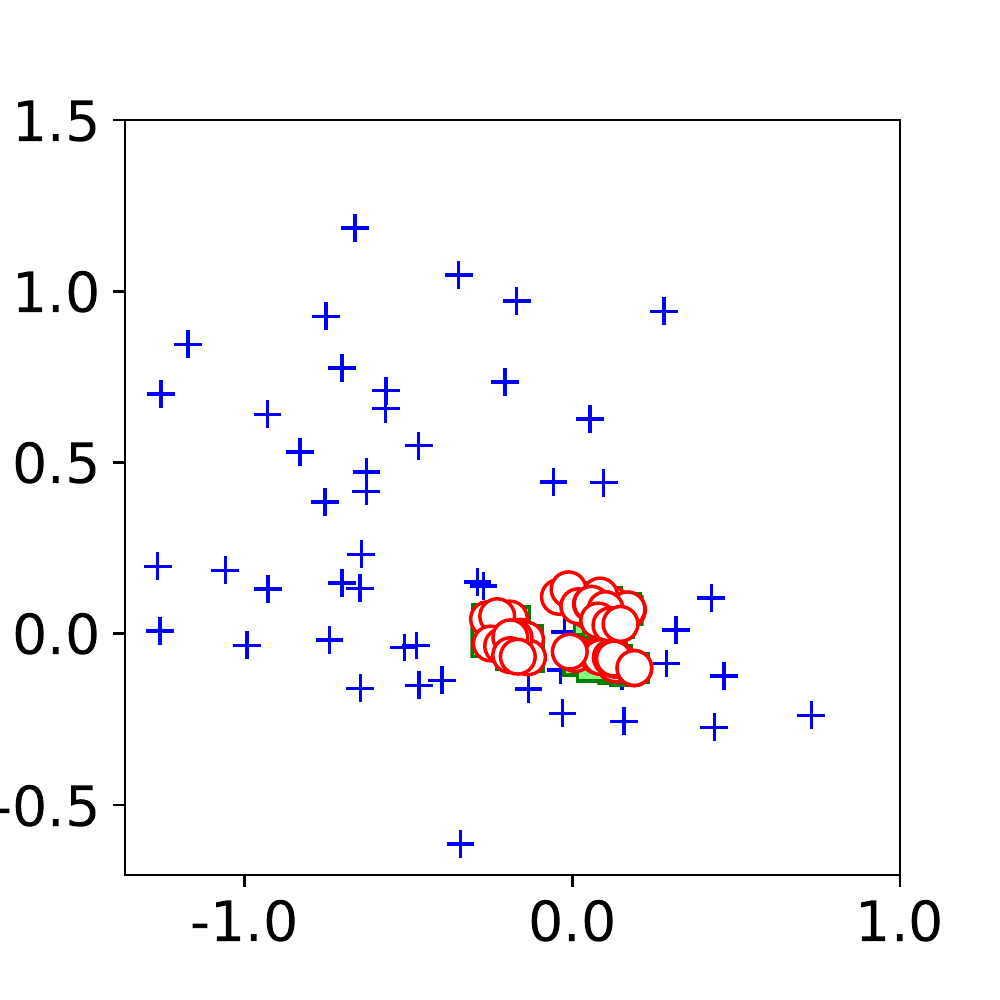}
	\subcaption{australian}
	\label{fig:australian_diversity}
	\end{minipage}
	\begin{minipage}[t]{0.26\textwidth}
	\includegraphics[width=0.93\textwidth]{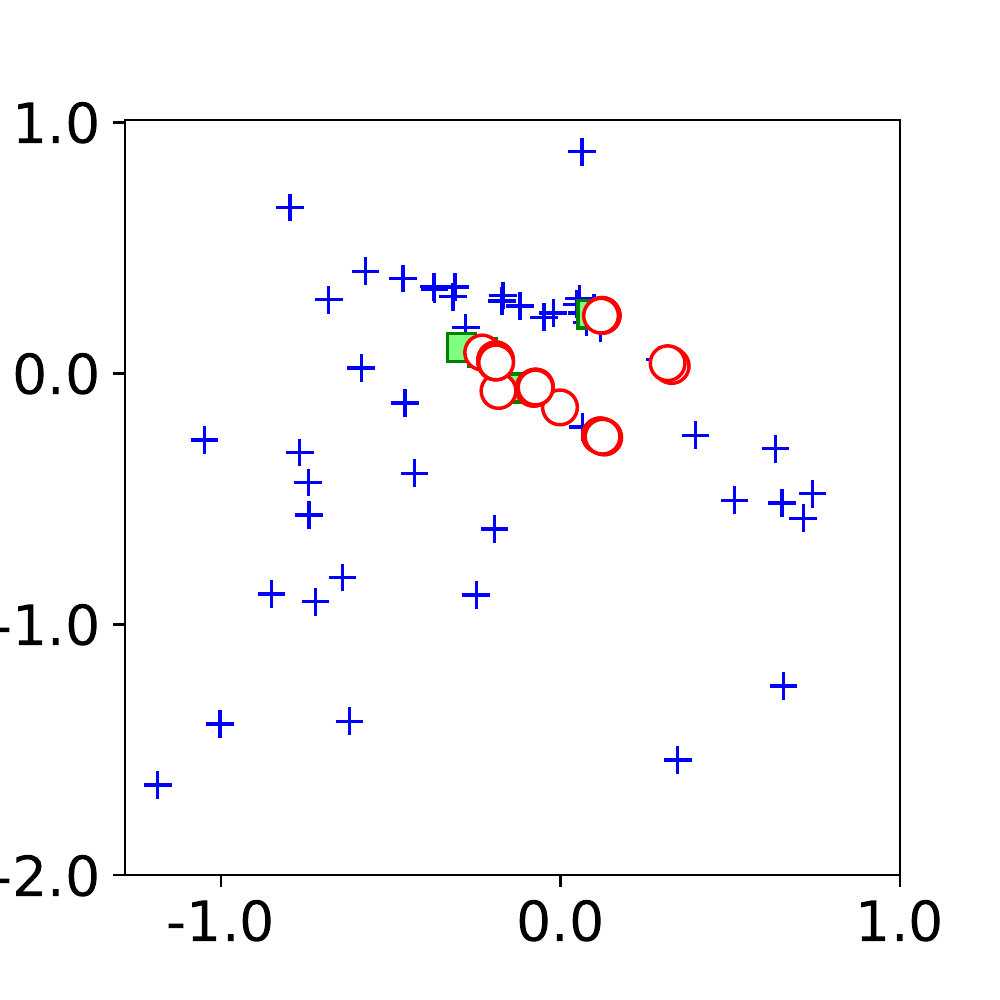}
	\subcaption{german.numer}
	\label{fig:german_numer_diversity}
    \end{minipage}
    \begin{minipage}[t]{0.10\textwidth}
    \begin{tikzpicture}
	    \useasboundingbox (3pt,0pt);
	    \draw [very thin] (3pt, 20pt) rectangle (73pt, 56pt);
		\node[circle, draw=red, fill=white, inner sep=0pt, minimum size=5pt, line width=1pt] (a) at (9pt, 48pt) {};
		\node[rectangle, draw={rgb:green,5;blue,2;red,2}, fill={rgb:green,2;white,1}, inner sep=0pt, minimum size=5pt, line width=1pt] (b) at (9pt, 38pt) {};
		\node[text=blue] (c) at (9pt, 28pt) {+};
		\node[right] at (15pt, 48pt) {Enumeration};
		\node[right] at (15pt, 38pt) {Heuristic};
		\node[right] at (15pt, 28pt) {Proposed};
	\end{tikzpicture}
	\end{minipage}\\
    \begin{minipage}[t]{0.26\textwidth}
	\includegraphics[width=0.93\textwidth]{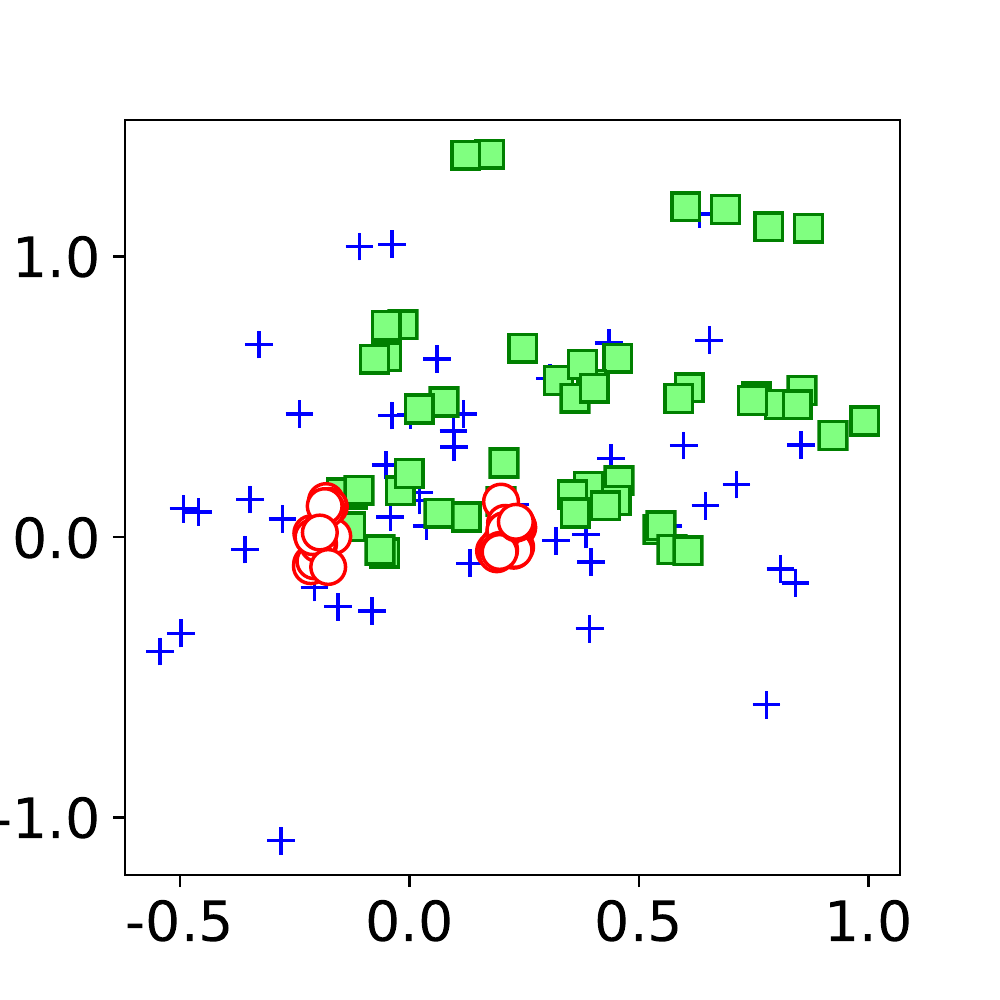}
	\subcaption{ionosphere}
	\label{fig:ionosphere_diversity}
	\end{minipage}
	\begin{minipage}[t]{0.26\textwidth}
	\includegraphics[width=0.93\textwidth]{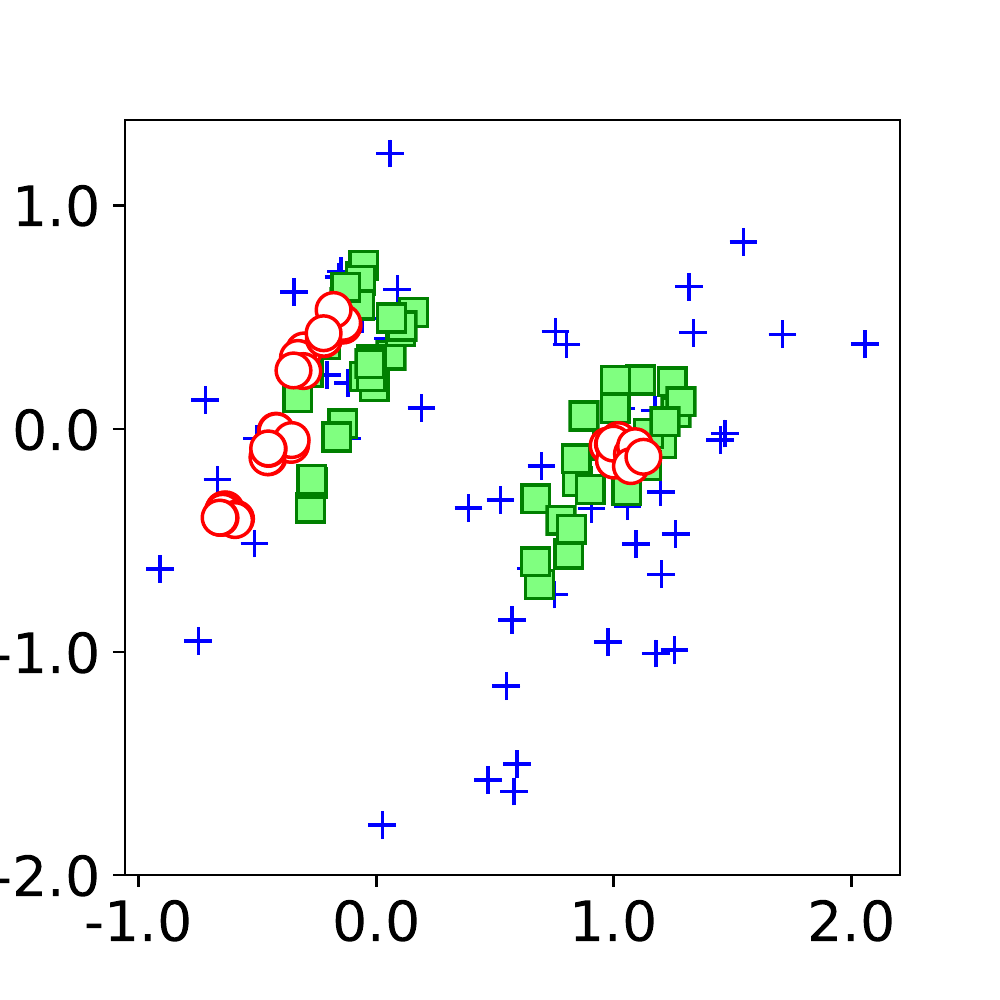}
	\subcaption{sonar}
	\label{fig:sonar_diversity}
	\end{minipage}
    \begin{minipage}[t]{0.26\textwidth}
	\includegraphics[width=0.93\textwidth]{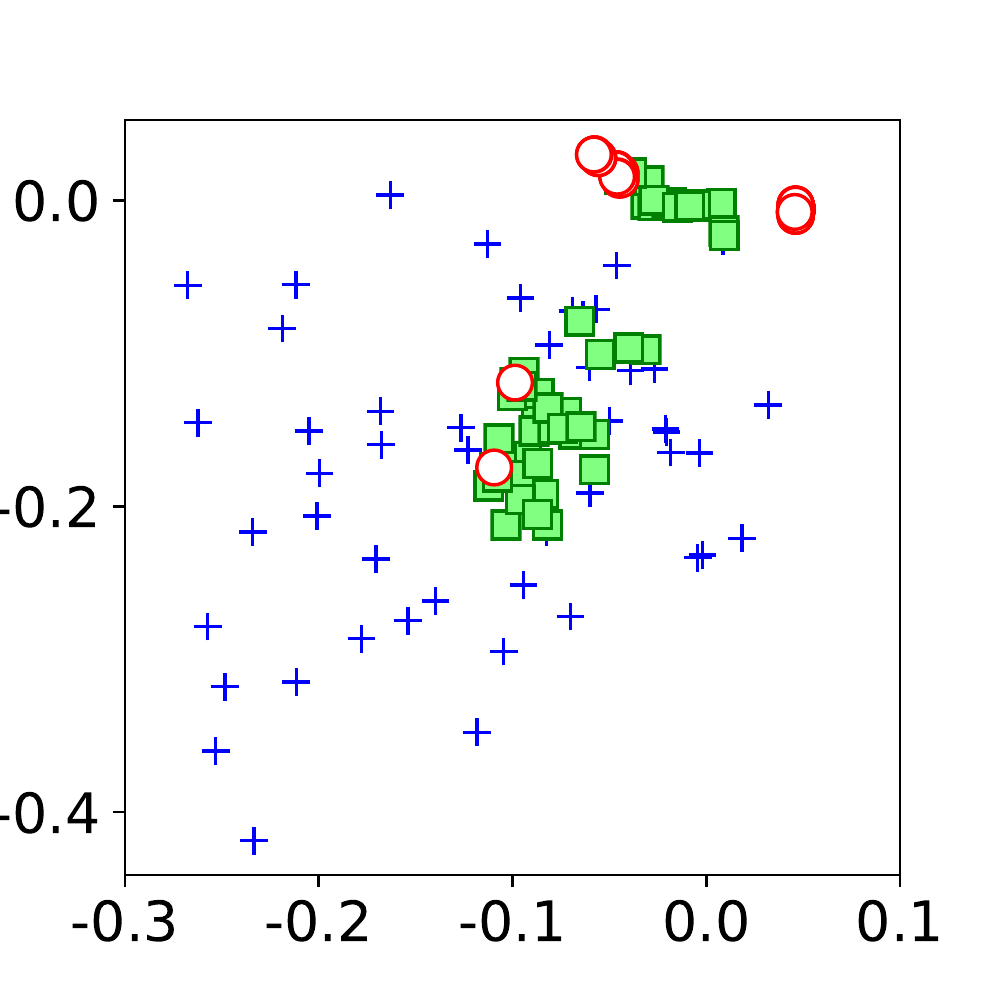}
	\subcaption{splice}
	\label{fig:splice_diversity}
	\end{minipage}
    \begin{minipage}[t]{0.10\textwidth}
    \hspace{1pt}
    \end{minipage}
	\caption{Found 50 solutions, shown in 2D using PCA}
	\label{fig:add_results}
\end{figure*}

\paragraph{Result}
With each method, we found 500 nearly optimal $\beta$, and summarized the result in \figurename~\ref{fig:20news_diversity}.
For the proposed method, we defined $B(\nu)$ by $\nu = 1.05L(\beta^*)$, and set the number of samplings $M$ to be 10,000.
To draw the figure, we used PCA and projected found solutions to the subspace where the variance of the solutions of Enumeration is maximum.
The figure shows the clear advantage of the proposed method in that it covers a large solution region compared to the other two baselines.
While the result indicates that Heuristic successfully improved the diversity of the found solutions compared to Enumeration, its diversity is still inferior to the ones of the proposed method.

We also note that the proposed method found 889 words in total within the 500 models.
This is contrastive to Enumeration and Heuristic where they found only 39 and 63 words, respectively, which is more than ten times less than the proposed method.
\tablename~\ref{tab:20news_words} shows some representative words found in 20 Newsgroups data.
As the word ``apple'' is strongly related to the documents in \texttt{mac.hardware}, it is found by all the methods.
However, although ``macs'' and ``macintosh'' are also relevant to \texttt{mac.hardware}, ``mac'' is overlooked by Enumeration, and ``macintosh'' is found only by the proposed method.
This result also suggests that the proposed method can induce a large diversity and it can avoid overlooking informative features.

We note that the Lasso global solution attained 81\% test accuracy, while the found 500 solutions attained from 77\% to 83\% test accuracies.
This result indicates that the proposed method could find solutions with almost equal qualities while inducing solution diversities.

\subsection{Results on Other Datasets}
\label{sec:results_other}

Here, we present the results on some of libsvm datasets\footnote{\url{https://www.csie.ntu.edu.tw/~cjlin/libsvmtools/datasets/}}: we used \texttt{cputime} for regression \eqref{eq:L}, and \texttt{australian}, \texttt{german.numer}, \texttt{ionosphere}, \texttt{sonar}, and \texttt{splice} for binary classification \eqref{eq:logreg}.
In the experiments, we searched for 50 near optimal solutions using the proposed method where we set $\nu = 1.05 L(\beta^*)$ and the number of samplings $M=1,000$.
We also enumerated 50 solutions using the two baselines, Enumeration and Heuristics.

The results are shown in \figurename~\ref{fig:add_results}.
For regression, we set $\rho = 0.1$, and for binary classification, we set $\rho = 0.01$, so that the solutions to be sufficiently sparse.
In the figures, similar to the results in Section~\ref{sec:diversity}, we have projected the solutions into two dimensional space using PCA.
The figures show the clear advantage of the proposed method in that it can find solutions with large diversities compared to the exhaustive enumerations.

%% file: conclusion.tex
\section{Conclusion}

In this study, we considered a convex hull approximation problem that seeks a small number of points such that their convex hull approximates the nearly optimal solution set to the Lasso regression problem.
We propose an algorithm to solve this problem. The algorithm first approximates the nearly optimal solution set by using the convex hull of sufficiently many points.
Then, it selects a few relevant points to approximate the convex hull.
The experimental results indicate that the proposed method can find diverse yet nearly optimal solutions efficiently.
